\documentclass[journal]{IEEEtran}
\usepackage{graphicx}
\usepackage{amsmath}
\usepackage{amssymb}
\usepackage{multirow}
\usepackage{subcaption}
\usepackage{hyperref}

\ifCLASSINFOpdf
\else
\fi
\begin{document}
\title{Multi-Residual Networks: Improving the Speed and Accuracy of Residual Networks}

\author{Masoud~Abdi,
        and~Saeid~Nahavandi,~\IEEEmembership{Senior~Member,~IEEE}% <-this % stops a space
\thanks{M. Abdi and S. Nahavandi are with the Institute for Intelligent Systems Research and Innovation (IISRI), Deakin University , Waurn Ponds,
VIC 3216, Australia. e-mail: mabdi@deakin.edu.au.
}
\thanks{This work has been submitted to the IEEE for possible publication. Copyright may be transferred without notice, after which this version may no longer be accessible.}% <-this % stops a space
}


\markboth{SUBMITTED FOR PUBLICATION, 2016}%
{Shell \MakeLowercase{\textit{et al.}}: Bare Demo of IEEEtran.cls for IEEE Journals}



\maketitle

\begin{abstract}
In this article, we take one step toward understanding the learning behavior of deep residual networks, and supporting the observation that deep residual networks behave like ensembles. We propose a new convolutional neural network architecture which builds upon the success of residual networks by explicitly exploiting the interpretation of very deep networks as an ensemble. The proposed multi-residual network increases the number of residual functions in the residual blocks. Our architecture generates models that are wider, rather than deeper, which significantly improves accuracy.  We show that our model achieves an error rate of 3.73\% and 19.45\% on CIFAR-10 and CIFAR-100 respectively, that outperforms almost all of the existing models. We also demonstrate that our model outperforms very deep residual networks by 0.22\% (top-1 error) on the full ImageNet 2012 classification dataset.  Additionally, inspired by the parallel structure of multi-residual networks, a model parallelism technique has been investigated. The model parallelism method distributes the computation of residual blocks among the processors, yielding up to 15\% computational complexity improvement.  %% competitive performance a test error rate of $3.92\%$ on CIFAR-10.
\end{abstract}


% \fi
\IEEEpeerreviewmaketitle

\section{Introduction}
\IEEEPARstart{C}{onvolutional}  neural networks \cite{lecun1989backpropagation} have contributed to a series of advances in tackling image recognition and visual understanding problems \cite{krizhevsky2012imagenet,sermanet2013overfeat,zeiler2014visualizing}. They have been applied in many areas of engineering and science \cite{wallach2015atomnet,mohan2014deep,kim2014convolutional}. Increasing the network depth is known to improve the model capabilities, which can be seen from AlexNet~\cite{krizhevsky2012imagenet} with 8 layers, VGG~\cite{simonyan2014very} with 19 layers, and GoogleNet~\cite{szegedy2015going} with 22 layers. However, increasing the depth can be challenging for the learning process because of the vanishing/exploding gradient problem~\cite{hochreiter1991untersuchungen,bengio1994learning}. Deep residual networks \cite{he2015deep} avoid this problem by using identity skip-connections, which help the gradient to flow back into many layers without vanishing. The identity skip-connections facilitate training of very deep networks up to thousands of layers that helped residual networks win five major image recognitions tasks in ILSVRC~2015 \cite{ILSVRC15} and Microsoft COCO~2015 \cite{lin2014microsoft} competitions.

However, an obvious drawback of residual networks is that every percentage of improvement requires significantly increasing the number of layers, which linearly increases the computational and memory costs \cite{he2015deep}. On CIFAR-10 image classification dataset, deep residual networks with 164-layers and 1001-layers reach a test error rate of $5.46\%$ and $4.92\%$ respectively, while the 1001-layer has six times more computational complexity than the 164-layer. On the other hand, wide residual networks \cite{zagoruyko2016wide} have 50 times fewer layers while outperforming the original residual networks. It seems that the power of residual networks is due to the identity skip-connections rather than extremely increasing the network depth.

Nevertheless, a recent study  supports that deep residual networks  act like ensembles of relatively shallow networks~\cite{veit2016residual}. This is achieved by showing the existence of exponential paths from the output layer to the input layer that gradient information can flow. Also, observations show that removing a layer from a residual network, during the test time, has a modest effect on its performance. Additionally, it shows that most of the gradient updates during optimization come from ensembles of relatively shallow depth. Moreover, residual networks do not resolve the vanishing gradient problem by preserving the gradient through the entire depth of the network. Instead, they avoid the problem by ensembling  exponential networks of different length. This raises the importance of \textit{multiplicity} that refers to the number of possible paths from the input layer to the output layer \cite{veit2016residual}.

Inspired by these observations, we introduce multi-residual networks (Multi-ResNet) which increase the multiplicity of the network, while keeping its depth fixed. This is achieved by increasing the  number of residual functions in each residual block. We then show that the accuracy of a shallow multi-residual network is similar to a deep 110-layer residual network. This supports that deep residual networks behave like ensembles instead of a single extremely deep network. Next, we examine the importance of \textit{effective range} which is the range of paths that significantly contribute towards gradient updates. 

We show that for a residual network deeper than a threshold $n_0$, increasing the number of residual functions leads to a better performance than increasing the network depth. This leads to a lower error rate for the multi-residual network with the same number of convolutional layers as the deeper residual network. Experiments on ImageNet, CIFAR-10, and CIFAR-100 datasets show that multi-residual networks improve the accuracy of deep residual networks and outperform almost all of the existing models. 

We demonstrate that a 101-layer Multi-ResNet with two residual functions in each block outperforms the top-1 accuracy rate of a 200-layer ResNet by $0.22\%$ on the ImageNet 2012 classification dataset \cite{ILSVRC15}. Also, using moderate data augmentation (flip/translation), multi-residual networks achieve an error rate of $4.35\%$ and $20.42\%$ on CIFAR-10 and CIFAR-100 receptively (based on five runs). This is \textbf{6\%} and \textbf{10\%} improvement compared to the residual networks with identity mappings \cite{he2016identity} with almost the same computational and memory complexity. The proposed multi-residual network achieves a test error rate of $3.73\%$ and $19.45\%$ on CIFAR-10 and CIFAR-100. 

Concurrent to our work, ResNeXt \cite{xie2016aggregated} and PolyNet \cite{zhang2016polynet} achieved second and third place in the ILSVRC~2016 classification task\footnote{http://image-net.org/challenges/LSVRC/2016/results}. Both models increase the number of residual functions in the residual blocks similar to our model, while PolyNet inserts higher order paths into the network as well. 

Eventually, a model parallelism technique has been explored to speed up the proposed multi-residual network. The model parallelism approach splits the calculation of each block between two GPUs, thus each GPU can simultaneously compute a portion of residual functions. This leads to the parallelization of the block, and consequently the network. The resulting network has been compared to a deeper residual network with the same number of convolutional layers that exploits data parallelism. Experimental results show that in addition to being more accurate, multi-residual networks can also be up to \textbf{15\%} faster.

In summary, the contributions of this research are:
\begin{itemize}
\item We take one step toward understanding deep residual networks and supporting that deep residual networks behave like ensembles of shallow networks, rather than a very deep network.

\item Through a series of experiments, we show the importance of the effective range in residual networks, which is the range of ensembles that significantly contribute toward gradient updates during optimization.

\item We introduce multi-residual networks that is shown to improve the classification accuracy of deep residual networks and many other state-of-the-art models.

\item We propose a model parallelism technique that is able to reduce the computational complexity of the multi-residual networks.

\end{itemize}

The rest of the paper is organized as follows. Section~\ref{sec2} details deep residual networks and other models capable of improving the original residual networks. The hypothesis that residual networks are exponential ensembles of relatively shallow networks is explained in Section~\ref{sec3}. The proposed multi-residual networks and the importance of the effective range are discussed in Section~\ref{sec4}.  Supporting experimental results are presented in Section~\ref{sec5}. Concluding remarks are provided in Section~\ref{sec6}. A pre-print version of this paper \cite{mine} is available at \url{https://arxiv.org/abs/1609.05672}, and the code to reproduce the results can be found at \url{https://github.com/masoudabd/multi-resnet}.

\section{Related Work}\label{sec2}

A residual block consists of a residual function $f$, and an identity skip-connection (see Figure \ref{fig:resblock}), where $f$ contains convolution, activation (ReLU) and batch normalization  \cite{ioffe2015batch} layers in a specific order. In the most recent residual network the order is normalization-ReLU-convolution which is known as the pre-activation model \cite{he2016identity}.

Deep residual networks contain many stacked residual blocks with $y = x + f(x)$, where $x$ and $y$ are the input and output of the block. Moreover, a deep residual network with the identity skip-connections \cite{he2016identity} can be represented as:

\begin{equation}\label{iresnet}
x_{l+1} = x_l+f_{l+1}(x_l)
\end{equation}
where $x_l$ is the input of $l^{\text{th}}$ residual block, and $f_l$ contains the weight layers. Additionally, Highway Networks \cite{srivastava2015training,srivastava2015highway} also employ parametrized skip-connections that are referred to as \textit{information highways}. The skip-connection parameters are learned during training, which control the amount of information that can pass through the skip-connections.

Residual networks with stochastic depth \cite{huang2016deep} use Bernoulli random variables to randomly disable the residual blocks during the training phase. This results in a shallower network at the training phase, while having a deeper network at the test phase. Deep residual networks with stochastic depth improve the accuracy of deep residual networks with constant depth. This is because of the reduction in the network depth which strengthens the back-propagated gradients of the earlier layers, and because of ensembling networks of different depths.

Swapout \cite{singh2016swapout} generalizes dropout \cite{srivastava2014dropout} and networks with stochastic depth \cite{huang2016deep} using $px+ qF(x)$, where $p$ and $q$ are two Bernoulli random variables. Swapout has the ability to sample from four network architectures $\{0, x, F(x), x+F(x)\}$, therefore having a larger domain for ensembles. Wide residual networks \cite{zagoruyko2016wide} increase the number of convolutional filters, and are able to yield a better performance than the original residual networks. This suggests that the power of residual networks originate in the residual connections, as opposed to extremely increasing the network depth. DenseNet \cite{huang2016densely} uses a dense connection pattern among the convolutional layers, where each layer is directly connected to all preceding layers.

\section{Deep Residual Networks behave like Ensembles}\label{sec3}

Deep residual networks \cite{he2015deep} are assumed to resolve the problem of vanishing gradients using identity skip-connections that facilitate training of deep networks up to 1202 layers. Nonetheless, recent studies support that deep residual networks do not resolve the vanishing gradient problem by preserving the gradient flow through the entire depth of the network. Instead, they avoid the problem simply by ensembling exponential networks together \cite{veit2016residual}.

Consider a residual network with three residual blocks, and let $x_0$ and $x_3$ be the input and output respectively, applying Equation \ref{iresnet} iteratively gives:
\begin{equation}\label{unrw}
\begin{split}
x_3 =&\ x_2 + f_3(x_2) \\
    =&\ \Big[ x_1 + f_2(x_1) \Big] + f_3(x_1+f_2(x_1)) \\
    =&\ \Big[ x_0 + f_1(x_0) + f_2(x_0+f_1(x_0)) \Big] \\
    +&\ f_3(x_0 + f_1(x_0) + f_2(x_0+f_1(x_0))) 
\end{split}
\end{equation}

\begin{figure}[!htb]
\centering
\includegraphics[width=0.95\linewidth]{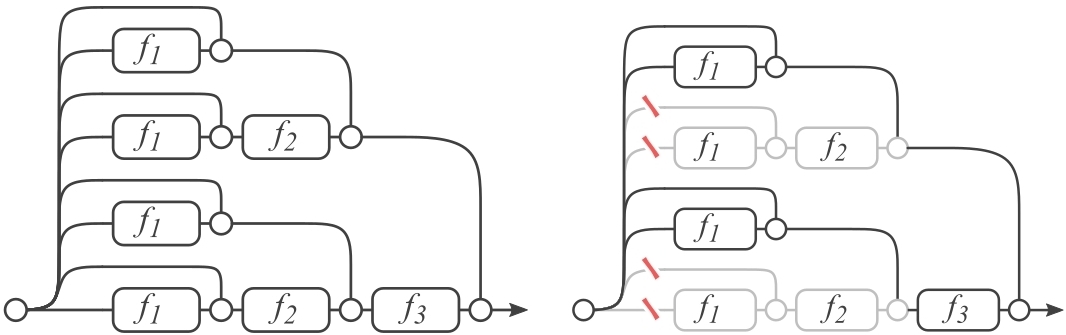}
%\captionsetup{width=0.4\textwidth}
\caption{(a) A residual network; (b) Deleting $f_2$ from a residual network~\protect\cite{veit2016residual}. It can be seen that residual networks have $2^n$ paths connecting the input to the output. Deleting a block from the residual network reduces the number of paths to $2^{n-1}$.}
\label{fig:unreview}
\end{figure}

A graphical view of Equation \ref{unrw} is presented in Figure \ref{fig:unreview}a. It is clear that data flow along the exponential paths from the input to the output layer. In other words, every path is a unique configuration that either computes a particular function $f_l(l=1,\dots,n)$ or skips it. Therefore, the total number of possible paths from the input to the output is $2^n$, where $n$ is the number of residual blocks. This term is referred to as the \textit{multiplicity} of the network.  Furthermore, a residual network can be viewed as a very large implicit ensemble of many networks with different length.

Deep residual networks are resilient to dropping and reordering the residual blocks during the test phase. More precisely, removing a single block from a 110-layer residual network, during the test phase, has a negligible effect on its performance. Whereas, removing a layer from the traditional network architectures, such as AlexNex\cite{krizhevsky2012imagenet} or VGGnet\cite{simonyan2014very}, dramatically hurts the performance of the models (test error more than $80\%$) \cite{veit2016residual}. This supports the existence of exponential paths from the input to the output layer. Moreover, removing a single residual block during the test phase reduces the number of paths from $2^n$ to $2^{n-1}$ (see Figure \ref{fig:unreview}b),

Additionally, shallow ensembles contribute significantly to the gradient updates during optimization . In other words, in a 110-layer residual network, most of the gradient updates come from paths with only 10-34 layers, and deeper paths do not have significant contribution towards  the gradient updates. These are called the \textit{effective paths}, which are relatively shallow compared to the network depth \cite{veit2016residual}.

In order to verify the claim pertaining to the shallow ensembles, one can see that individual paths in a deep residual network have a binomial distribution, where the number of paths with length $k$ is $\binom{n}{k} = \frac{n!}{k!(n-k)!}$. On the other hand, it has been known that the gradient magnitude, during back-propagation,  decreases exponentially with the number of functions it goes through \cite{hochreiter1991untersuchungen,bengio1994learning}. Therefore, the total gradient magnitude contributed by paths of each length can be calculated by multiplying the number of paths with that length, and the expected gradient magnitude of the paths with the same length~\cite{veit2016residual}.

Accordingly, a residual network trained  with only effective paths has a comparable performance with the full residual network \cite{veit2016residual}. This is achieved by randomly sampling a subset of residual blocks for each mini-batch, and forcing the computation to flow through the selected blocks only. In this case the network can only see the effective paths that are relatively shallow, and no long path is used.

\section{Multi-Residual Networks}\label{sec4}

Based on the aforementioned observations, we propose \textit{multi-residual networks} that aim to increase the multiplicity of the residual network, while keeping the depth fixed. The multi-residual network employs multiple residual functions, $f^i$, instead of one function for each residual block (see Figure~\ref{fig:resblock}). As such, a deep multi-residual network with $k$ functions has:

\begin{figure}
\centering
\includegraphics[width=0.95\linewidth]{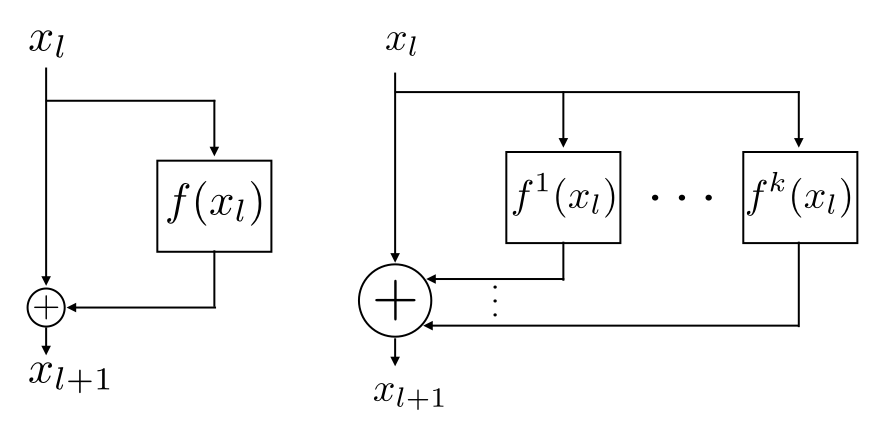}
\caption{A residual block (left) versus a multi-residual block (right).}
\label{fig:resblock}
\end{figure}

\begin{equation}\label{mwresnet}
x_{l+1} = x_l+ f_{l+1}^1(x_l) + f_{l+1}^2 (x_l) + \dots + f_{l+1}^k(x_l)
\end{equation}
where $f^i_l$ is the $i^{th}$ function of the $l^{th}$ residual block. Expanding Equation~\ref{mwresnet} for $k=2$ functions and three multi-residual blocks gives:

\begin{equation}\label{mwexpanded}
\begin{split}
x_3 =& x_2 + f_3^1(x_2) + f_3^2(x_2) \\
    =& \Big[ x_1 + f_2^1(x_1) + f_2^2(x_1)\Big] + \Big[f_3^1(x_1 + f_2^1(x_1) + f_2^2(x_1))\Big] \\
    +& \Big[f_3^2(x_1 + f_2^1(x_1) + f_2^2(x_1))\Big]\\
    =& \Big[ x_0 + f_1^1(x_0) + f_1^2(x_0)+ f_2^1(x_0+f_1^1(x_0)+f_1^2(x_0))\\
    & +f_2^2(x_0+f_1^1(x_0)+f_1^2(x_0))\Big] \\
    +& \Big[f_3^1(x_0 + f_1^1(x_0)+ f_1^2(x_0)+f_2^1(x_0 + f_1^1(x_0)+ f_1^2(x_0)) \\
    &+ f_2^2(x_0 + f_1^1(x_0)+ f_1^2(x_0))) \Big]  \\
    +& \Big[f_3^2(x_0 + f_1^1(x_0)+ f_1^2(x_0)+f_2^1(x_0 + f_1^1(x_0)+ f_1^2(x_0)) \\
    &+ f_2^2(x_0 + f_1^1(x_0)+ f_1^2(x_0))) \Big] 
\end{split}
\end{equation}
It can be seen that the number of terms in Equation~\ref{mwexpanded} is exponentially more than the number of terms in Equation~\ref{unrw}. Specifically, in a multi-residual block with $k=2$ residual functions, the gradient flow has four possible paths: (1)~skipping both $f^1$ and $f^2$, (2)~skipping $f^1$ and performing $f^2$, (3)~skipping $f^2$ and performing $f^1$, (4)~performing both $f^1$ and $f^2$. Therefore, the multiplicity of the multi-residual network with two residual functions is $4^n$. In other words, the multiplicity of a multi-residual network with $k$ residual functions and $n$ multi-residual blocks is $2^{kn}$. This is because every function can be either computed or otherwise, giving a multiplicity of $2^k$ for a block, and a total multiplicity of $2^{kn}$ for the multi-residual network.

\subsection{Residual Networks Behave Like Ensembles}

Based on multi-residual networks, we show that residual networks behave like ensembles. A shallow multi-residual network with the same number of parameters as a 110-layer residual network is able to achieve the accuracy of the residual network. This supports the hypothesis that residual networks behave like exponential ensembles of shallow networks, rather than a single deep network.

\begin{table}[!htb]
\centering
\begin{tabular}{||c c c c c ||} 
\hline
method & depth & k & \#params & CIFAR-10(\%)  \\ 
\hline\hline
\multirow{1}{7.5em}{resnet\cite{he2015deep}} 
& 110 & 1& 1.7M & 6.61 \\ 
\hline
\multirow{1}{7.5em}{pre-resnet\cite{he2016identity}} 
& 110 & 1& 1.7M & 6.37 \\ 
\hline
\multirow{2}{7.5em}{multi-resnet [ours]} 
& 8 & 23 & 1.7M & 7.37\\ 
& 14 & 10 & 1.7M & 6.42 \\ 

\hline
\end{tabular}
%\captionsetup{width=0.4\textwidth}
\caption{Classification error on CIFAR-10 test set. A shallow multi-residual network is able to approximate the accuracy of a 110-layer residual network.}
\label{table:1}
\end{table}

\begin{figure*}[!t]
    \centering
    \begin{subfigure}[b]{0.4\textwidth}
        \includegraphics[width=\textwidth]{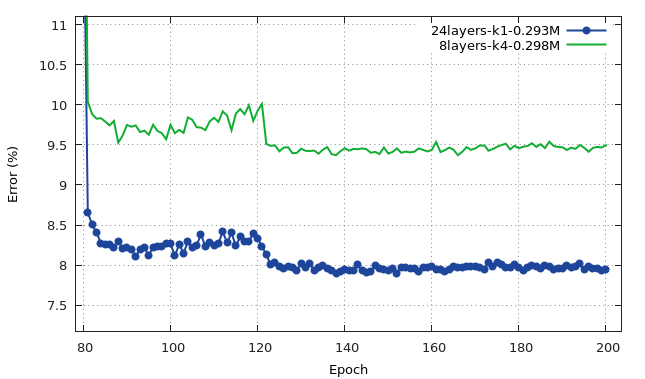}
        \caption{}
        \label{fig:a}
    \end{subfigure}
    ~ %add desired spacing between images, e. g. ~, \quad, \qquad, \hfill etc. 
    \begin{subfigure}[b]{0.4\textwidth}
        \includegraphics[width=\textwidth]{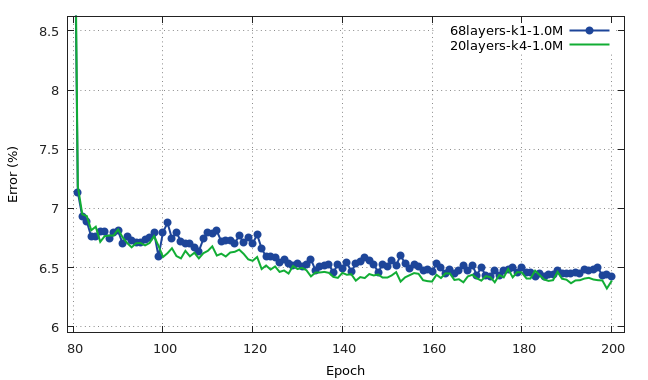}
        \caption{}
        \label{fig:b}
    \end{subfigure}
    ~ %add desired spacing between images, e. g. ~, \quad, \qquad, \hfill etc. 
    \begin{subfigure}[b]{0.4\textwidth}
        \includegraphics[width=\textwidth]{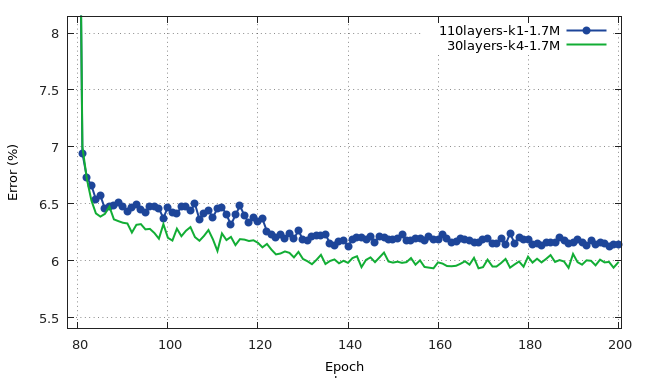}
        \caption{}
        \label{fig:c}
    \end{subfigure}
    \caption{Comparing residual network and the proposed multi-residual network on CIFAR-10 test set to show the effective range phenomena. Each curve is mean over 5 runs. (a) This is the situation that the network depth $<n_0$ in which the multi-residual network performs worse than the original residual network; (b) Both networks have a comparable performance; (c) The proposed multi-residual network outperforms the original residual network.}\label{fig:tradeoff}
\end{figure*}

A Multi-ResNet with the depth of 8 and $k=23$ residual functions, and a Multi-ResNet with the depth of 14 and $k=10$ residual functions are trained. Both networks have roughly the same number of parameters, which is the same as those in the 110-layer residual network. The networks are trained with the same hyper-parameters and training policy as in \cite{he2015deep}. Table~\ref{table:1} summarizes the test errors on CIFAR-10. It can be seen that the classification accuracy of the shallow multi-residual network with 14-layer depth almost reaches that of the 110-layer residual network.

\subsection{The Effective Range}

Based on the observation that residual networks behave like ensembles of shallow networks, a question is posed: \textbf{what is the relationship between the range of the effective paths and the depth of the residual network?} More precisely, what is the relationship between the effective range of a residual network with $n$ residual blocks and that of a residual network with $cn$ residual blocks, where $c$ is a constant number?

We hypothesize that this relationship is \textbf{not linear}. This implies that if the effective range of a residual network with $n$ blocks is $[a,b]$, the effective range of a residual network with $cn$ blocks is not $[ca,cb]$. Instead, it is shifted and/or scaled toward shallower networks. This is because of the exponential reduction in the gradient magnitude \cite{veit2016residual,bengio1994learning}. Eventually, the upper bound of the effective range is lower than $cb$. This could be a potential reason for the problem that every percentage of improvement in deep residual networks requires significantly increasing the number of layers.  %We note that this effect happens only when the depth $n$ is greater than a large enough $n_0$. 

\subsection{Residual Networks versus Multi-Residual Networks}\label{vs}

Consider a residual network $R$ with $n$ residual blocks, and let $c$ be a constant integer. We would like to construct two residual networks by: (1) increasing the number of residual blocks to $cn$, which results in a residual network with $c$ times depth of~$R$ (excluding the first and last layers), (2) retaining the same depth while increasing the number of residual functions by~$c$. The number of parameters of the subsequent networks are roughly the same. One can also see that the multiplicity of both networks are $2^{cn}$, but \textbf{how about the effective range of (1) and (2)}?

As discussed in the previous part, the effective range of (1) does not increase linearly, whereas the effective range of (2)  increases linearly due to the increase in the residual functions. This is owing to the increase in the number of  paths of each length, which is a consequence of  changing the binomial distribution to a multinomial distribution. Note that this analysis holds true for $n\geq n_0$, where $n_0$ is a threshold; otherwise the power of the network depth is clear both in theory \cite{haastad1991power,haastad1987computational,eldan2015power} and in practice \cite{krizhevsky2012imagenet,simonyan2014very,szegedy2015going}.

\section{Experimental Results}\label{sec5}

To support our analyses and show the effectiveness of the proposed multi-residual networks, a series of experiments has been conducted on CIFAR-10 and CIFAR-100 datasets.  Both datasets contain 50,000 training samples and 10,000 test samples of $32\times 32$ color images, with 10 (CIFAR-10) and 100 (CIFAR-100) different categories.   We have used \textit{"moderate data augmentation"} (flip/translation) as in \cite{he2016identity}, and training is done using stochastic gradient descent for 200 epochs with a weight decay of $10^{-4}$ and momentum of 0.9 \cite{he2015deep}. The network weights have been  initialized as in \cite{he2015delving}. %The learning rate starts with 0.1 and divides by 10 at epochs 80 and 120. The code is available at: \url{https://github.com/masoudabd/multi-resnet}

\subsection{The Effective Range Phenomena}\label{effph}

Consider a pre-activation version of the residual network with the basic-blocks \cite{he2016identity}. Three pairs of residual network and multi-residual network are trained. The residual network is $k$ times deeper than the corresponding multi-residual network (excluding the first and last layers). On the other hand, the multi-residual network computes $k$ residual functions. A residual block might be removed to compensate the difference in the number of parameters to form a fair comparison between the pairs. The median of five runs with mean$\pm$std in the parentheses are reported in Table \ref{table:2}. Test error curves are also depicted in Figure \ref{fig:tradeoff}, where each curve is the mean of five runs. All networks are trained with the same hyper-parameters and training policy with a mini-batch size of 128.

\begin{table}[!htb]
\centering
\begin{tabular}{||c c c c c ||} 
\hline
method & depth & k & \#params & CIFAR-10(\%)   \\ 
\hline\hline
\multirow{3}{5em}{pre-resnet \cite{he2016identity}} 
& 24 & 1& 0.29M & 7.75 \tiny(7.76$\pm$0.13) \\ 
& 68 & 1& 1.0M & 6.27 \tiny(6.33$\pm$0.24) \\ 
& 110 & 1& 1.7M & 6.02 \tiny(6.02$\pm$0.11)  \\ 
\hline
\multirow{3}{5em}{multi-resnet [ours]} 
& 8 & 4& 0.29M & 9.28 \tiny(9.28$\pm$0.07) \\ 
& 20 & 4& 1.0M & 6.31 \tiny(6.29$\pm$0.22) \\ \
& 30 & 4& 1.7M & 5.89 \tiny(5.85$\pm$0.12) \\ 

\hline

\end{tabular}
%\captionsetup{width=0.4\textwidth}
\caption{CIFAR-10 test errors of the multi-residual networks and the original residual networks, where $k$ is the number of functions. The results are in the form of $median$ with $mean\pm std$ in parentheses from five runs.}
\label{table:2}
\end{table}

The Multi-ResNet with 8-layers depth has a test error rate of $9.28\%$, while the original ResNet with 24 layers, and roughly the same number of parameters, has an error rate of $7.75\%$. This is the scenario whereby the network depth is too shallow $(depth<n_0)$, and the multi-residual network performs worse than the residual network (see Figure~ \ref{fig:a}). On the contrary, the Multi-ResNet with 20-layers depth achieves $6.31\%$ error rate, which is statistically no different than  $6.27\%$ for the 68-layer ResNet.  Test curves (Figure~\ref{fig:b}) also show that both networks have a comparable performance.

\begin{table*}[!htp]
\centering
\begin{tabular}{||c c c c|c |c||} 
\hline
\multicolumn{4}{||c|}{method} & CIFAR-10(\%) & CIFAR-100(\%)\\
\hline
\hline
\multicolumn{4}{||c|}{NIN\cite{lin2013network}} & 8.81 & 35.68\\
\multicolumn{4}{||c|}{DSN\cite{lee2015deeply}} & 8.22 & 34.57\\
\multicolumn{4}{||c|}{FitNet\cite{romero2014fitnets}} & 8.39 & 35.04\\
\multicolumn{4}{||c|}{Highway\cite{srivastava2015training}} & 7.72 & 32.39\\
\multicolumn{4}{||c|}{All-CNN\cite{springenberg2014striving}} & 7.25 & 33.71\\
\multicolumn{4}{||c|}{ELU\cite{clevert2015fast}} & 6.55 & 24.28\\

\hline
\hline
method & depth & k,(w) & \#parameters & &  \\
\hline
\hline

\multirow{2}{8em}{resnet\cite{he2015deep}} 
& 110 & 1 & 1.7M & 6.43\tiny(6.61$\pm$0.16) & 25.16 \\ 
& 1202 & 1 & 19.4M & 7.93 & 27.82 \\ 
\hline
\multirow{3}{8em}{pre-resnet\cite{he2016identity}} 
& 110 & 1& 1.7M & 6.37 & - \\ 
& 164 & 1& 1.7M & 5.46 & 24.33 \\ 
%& 1001 & 1& 10.2M & 4.92\tiny(4.89$\pm$0.14) & 22.71\tiny(22.68$\pm$0.22) \\ 
& 1001 & 1& 10.2M & 4.62\tiny(4.69$\pm$0.20)$^\dagger$ & 22.71\tiny(22.68$\pm$0.22) \\ 
\hline
\multirow{2}{8em}{stoch-depth\cite{huang2016deep}} 
& 110 & 1 & 1.7M & 5.25 & 24.58 \\ 
& 1001 & 1 & 10.2M & 4.91 & - \\ 
\hline
\multirow{2}{8em}{swapout\cite{singh2016swapout}} 
& 20 & 1,(2) & 1.1M & 6.58 & 25.86 \\ 

& 32 & 1,(4) & 7.43M & 4.76 & 22.72 \\ 
\hline
\multirow{3}{8em}{wide-resnet\cite{zagoruyko2016wide}} 
& 40 & 1,(4) & 8.7M & 4.97 & 22.89 \\ 
& 16 & 1,(8) & 11.0M & 4.81 & 22.07 \\ 
& 28 & 1,(10) & 36.5M & 4.17 & 20.50 \\ 

\hline
\multirow{2}{8em}{DenseNet\cite{huang2016densely}$^\dagger$} 
& 100 & 1 & 7.0M & 4.10 & 20.20 \\ 
& 100 & 1 & 27.2M & 3.74 & \textbf{19.25} \\ 
\hline
\multirow{4}{8em}{multi-resnet [ours]$^\dagger$} 
%& 200 & 5 & 10.2M & 4.58\tiny(4.57$\pm$0.09) & 22.66\tiny(22.55$\pm$0.34) \\ 
& 200 & 5 & 10.2M & \textbf{4.35}\tiny(4.36$\pm$0.04)& \textbf{20.42}\tiny(20.44$\pm$0.15)\\
& 398 & 5 & 20.4M & 3.92 & 20.59 \\
& 26 & 2,(10) & 72M & 3.96 & \textbf{19.45} \\ 
& 26 & 4,(10) &  145M& \textbf{3.73} & 19.60 \\

\hline

\end{tabular}
\caption{Comparison of test error rates on CIFAR-10 and CIFAR-100. The results in the form of $median(mean\pm std)$ are based on five runs, while others are based on one run. All results are obtained with a mini-batch size of~128, except $^\dagger$ with a mini-batch size of 64. The number of residual functions in each residual block is denoted as k, and (w) is the widening factor for wider models.}
\label{table:cifar}
\end{table*}

Eventually, a 30-layer deep Multi-ResNet achieves $5.89\%$ error rate. This is slightly better than the 110-layer ResNet that have the error of $6.02\%$ ($6.37\%$~in~\cite{he2016identity}). Figure~\ref{fig:c} also clearly shows that the multi-residual network performance is superior to that of the original residual network. It can be seen that although each pair have almost the same number of parameters and computational complexity, they act very differently. These results support the hypothesis pertaining to the effective range.

In the previous section, we argue that multi-residual network is able to improve classification accuracy of the residual network when the network is deeper than a threshold $n_0$. This effect can be seen in Figure~\ref{fig:tradeoff}. Based on the observations in Table~\ref{table:2}, for this particular dataset and network/block architecture, the threshold $n_0$ is approximately 20. Furthermore, by increasing the number of functions, better accuracy can be obtained. However,  a trade-off has been observed between the network depth and the number of function. This means that, one might need to choose a suitable number of residual functions, and depth to achieve the best performance.

\subsection{CIFAR Experiments}%Improving the Accuracy of Existing Models}

Table \ref{table:cifar} shows the results of  multi-residual networks along with those from the original residual networks  and other state-of-the-art models. The networks with $6n+2$ layers use the basic block with two $3\times 3$ convolutional layers, and the networks with $9n+2$ layers use the bottleneck block architecture, which has a single $3\times 3$ convolutional layer surrounded by two $1\times 1$ convolutional layers \cite{he2015deep}. We also trained wider  \cite{zagoruyko2016wide} versions of Multi-ResNet and show that it achieves state-of-the-art performance. One can see that the proposed multi-residual network outperforms almost all of the existing models on CIFAR-10 and CIFAR-100 with the test error rate of $3.73\%$ and $19.45\%$ respectively.

%These models include the original residual networks \cite{he2015deep}, pre-activation residual networks with identity mappings \cite{he2016identity}, residual networks with stochastic depth \cite{huang2016deep}, swapout \cite{singh2016swapout}, and wide residual networks \cite{zagoruyko2016wide}. 

\textbf{Complexity of the proposed model.} Increasing the number of residual functions by $k$ increases the number of parameters by a factor of $k$, and  the computational complexity of the multi-residual network also increases linearly with the number of residual functions. This results in the memory and computational complexity similar to those of the original residual networks with the same number of convolutional layers \cite{he2016identity}.

%It is necessary to mention that all of our results are obtained using the same number of convolutional filters as the original residual networks~\cite{he2015deep}. Therefore, it is possible that a wider version of multi-residual networks attain a better performance than those in Table~\ref{table:4}.

\subsection{ImageNet Experiments}

We also perform experiments on the ImageNet 2012 classification dataset~\cite{ILSVRC15}. ImageNet is a dataset containing around 1.28 million training images from 1000 categories of objects that is largely used in computer vision applications. All trainings, in this section, are done using stochastic gradient descent up to 90 epochs. The hyper parameters described earlier are used, excluding the learning rate which is divided by 10 every 30 iteration. The networks are trained and tested on $224\times 224$ crops using scale and aspect ratio augmentation \cite{he2016identity,szegedy2015going}.% with the size of 

\begin{table}[!htb]
\centering
\begin{tabular}{||c c c c c ||} 
\hline
method & depth & k & Top-1(\%) & 10-Crop(\%)   \\ 
\hline\hline
\multirow{2}{4.9em}{pre-resnet \cite{he2016identity}} 
& 34 & 1& 26.73 & 24.77 \\ 
%& 101 & 1& 22.44 & 6.21 \\ 
& 200 & 1& 21.66 & 20.15 \\ 
\hline
\multirow{2}{4.9em}{multi-resnet [ours]} 
& 18 & 2& 27.39 & 25.61 \\ 
%& 50 & 2& TBA &  TBA \\
& 101 & 2& \textbf{21.53} & \textbf{19.93}  \\ 
\hline

\end{tabular}
%\captionsetup{width=0.4\textwidth}
\caption{Top-1 error rate comparison of deep residual networks and multi-residual networks on ILSVRC 2012 validation set. Multi-residual network outperforms deep residual networks.}
\label{table:img}
\end{table}

Table \ref{table:img} verifies that multi-residual network outperforms a deep residual network with the same number of convolutional layer, as long as the networks are deeper than a threshold. Specifically, the 101-layer Multi-ResNet with two residual functions outperforms the 200-layer ResNet by $0.13\%$ top-1 error rate with the same computational complexity. By testing on multiple crops, the Multi-ResNet outperforms the residual network with $0.22\%$.% Due to our limited resources, we could only train a 101-layer Multi-ResNet completely, however, it is highly expected that a deeper Multi-ResNet significantly improve these results.

Concurrently, ResNeXt \cite{xie2016aggregated} and PolyNet \cite{zhang2016polynet} obtained  second and third place in the ILSVR 2016 classification task with $3.03\%$ and $3.04\%$ top-5 error rate respectively. They are similar to our network architecture in the sense that they both increase the number of functions in the residual blocks. PolyNet also exploits second order paths that compute two functions sequentially in the same block.

\subsection{Toward Model Parallelism}
Although deep residual networks are extremely accurate, their computational complexity is a serious bottleneck to their performance. On the other hand, by simply implementing multi-residual networks, one does not make use of the increase in network width and the reduction in network depth. This is because eventually the residual functions in each residual block are computed and added in a sequential manner. Moreover, the parallel structure of multi-residual networks inspired us to examine the effects of model parallelism as opposed to the more commonly used data parallelism.

\begin{figure}[!htb]
\centering
\includegraphics[width=0.95\linewidth]{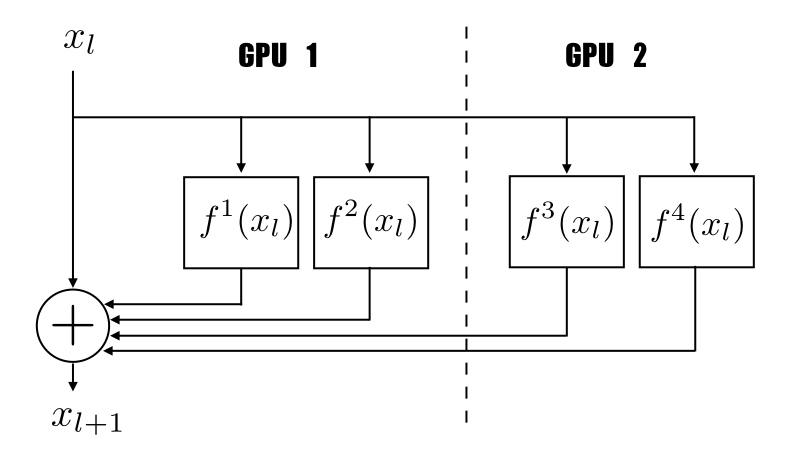}
\caption{Model parallelization of a multi-residual block with four residual functions on two GPUs.}
\label{fig:modelparallel}
\end{figure}

Data parallelism splits the data samples among the available GPUs and every GPU computes the same network on its portion of data (Single Instruction Multiple Data), and sends the results back to the main GPU to perform the optimization step. On the contrary, model parallelism splits the model among the desired GPUs and each GPU computes a different part of the model on the same data (Multiple Instruction Single Data)  \cite{dean2012large,krizhevsky2014one}. More precisely, for every multi-residual block with $k$ residual functions, we split the model between two GPUs and each GPU calculates $k/2$ of the residual functions in both forward and backward passes (see Figure~\ref{fig:modelparallel}). The results are then combined on the first GPU to perform the optimization step. Furthermore, the parallelization of each block is believed to reduce the total computational cost of the network.

\begin{table*}[!htb]
\centering
\begin{tabular}{||c c c|| c c c ||c ||c||} 
\hline
\multicolumn{3}{||c||}{pre-resnet \cite{he2016identity}} &
\multicolumn{3}{c||}{multi-resnet [ours]} &
\multicolumn{1}{c||}{ mini-batch  } &
\multicolumn{1}{c||}{ speed up } \\

\cline{1-6}
depth & k & Time  &depth & k & Time & size & \\  
\hline
\hline

218 & 1& 413ms & 110 & 2& 462ms &128 & -\\ 

434 & 1& 838ms & 110 & 4& 804ms &128 & 4\%\\ 

650 & 1& 1284ms & 110 & 6& 1158ms &128 & 10\%\\ 
 
%866 & 1& 540ms & 110 & 8& 450ms &128 & 16\%\\ 

\hline

218 & 1& 137ms & 110 & 2& 136ms &32 & 1\%\\ 

434 & 1& 273ms & 110 & 4& 238ms &32 & 13\%\\ 

650 & 1& 402ms & 110 & 6& 341ms &32 & 15\%\\ 
 
%866 & 1& 540ms & 110 & 8& 450ms &32 & 16\%\\ 

\hline

\end{tabular}
%\captionsetup{width=0.4\textwidth}
\caption{Computational time comparison between multi-residual networks and deep residual networks. The captured times are the time for a single SGD step using two GPUs. Multi-residual networks use model parallelism and residual networks exploit data parallelism.}
\label{table:5}
\end{table*}

Using the proposed model parallelism, we compare the computational complexity of the multi-residual network with a similar deep residual network that exploits data parallelism. All experiments are done using Nvidia Tesla K80 GPUs which consists of two sub GPUs connected with a PCI-Express (Gen3) link. This link is capable of transferring data up to 16 GB/s. The elapsed time for a single stochastic gradient descent step including the forward pass, backward pass and parameter update are shown in Table~\ref{table:5}.

In the proposed model parallelism, the inputs and outputs of blocks must be  transferred between the GPUs, which occupies most of the computational time. While, in data parallelism, every GPU performs a single forward and backward step independent of others. Nevertheless, Table~\ref{table:5} demonstrates that the multi-residual network with model parallelism still has less computational complexity than the corresponding residual network. However, this might not be true in some network architectures because of the communication overhead.

Interestingly, this effect amplifies when the number of data samples on each GPU  become less than 32, owing to the fact that threads on the current Nvidia GPUs are dispatched in the arrays of 32 threads (called wrap). Therefore, the computational power of GPU is wasted when there are only 16 samples on the GPU. This is sometimes the case in large-scale training, where one has to reduce the batch size in order to fit a larger network in the GPU memory. Also sometimes smaller mini-batch size obtains better accuracy \cite{he2016identity}.

Consequently, in order to exploit the advantages of both model and data parallelism, one can utilize a hybrid parallelism. As a result, the hybrid parallelism performs data parallelism among four (K80) GPUs and each  GPU performs model parallelism internally between the two sub GPUs. This offers up to 15\% computational complexity improvement with respect to the deeper residual network.%[NOT SURE] Accordingly, one limitation of the hybrid approach is that the required memory will increase with respect to the complete data parallelism. Moreover, techniques such as sharing gradient matrix of the block can compensate for the limitaiton of the memory requirement.

\section{Conclusions}\label{sec6}

Experiments in this article support the hypothesis that deep residual networks behave like ensembles, rather than a single extremely deep network. Based on a series of analyses and observations, multi-residual networks are introduced. Multi-residual networks exploit multiple functions for the residual blocks which leads to networks that are wider, rather than deeper. The proposed multi-residual network is capable of enhancing classification accuracy of the original residual network and almost all of the existing models on ImageNet, CIFAR-10, and CIFAR-100 datasets. Finally, a model parallelism technique has been investigated to reduce the computational cost of multi-residual networks. By splitting the computation of the multi-residual blocks among processors, the network is able to perform the computation faster.

% or


\section*{Acknowledgment}

The authors would like to thank the National Computational Infrastructure for providing us with high-performance computational resources. We thank Hamid Abdi and Chee Peng Lim for their useful discussions.

\ifCLASSOPTIONcaptionsoff
  \newpage
\fi

{\small
\bibliographystyle{abbrv}
\bibliography{mresnet}
}

\end{document}